\documentclass[10pt,twocolumn,letterpaper]{article}

\usepackage[pagenumbers]{cvpr} 
\usepackage{xcolor}

\usepackage{algorithm}
\usepackage{algpseudocode}

\usepackage{amsmath}
\usepackage[hypcap=false]{caption}
\definecolor{citecolor}{HTML}{2980b9}
\definecolor{linkcolor}{HTML}{c0392b}
\definecolor{mycolor_blue}{HTML}{E7EFFA}
\definecolor{mycolor_green}{HTML}{E6F8E0}
\definecolor{mycolor_gray}{HTML}{ECECEC}

\definecolor{cvprblue}{rgb}{0.21,0.49,0.74}
\usepackage[pagebackref,breaklinks,colorlinks,allcolors=cvprblue]{hyperref}
\usepackage{float}
\usepackage{multirow}
\def\paperID{11349} 
\def\confName{CVPR}
\def\confYear{2025}

\title{Towards Transformer-Based Aligned Generation \\ with Self-Coherence Guidance}

\author{Shulei Wang$^{1*}$ \quad Wang Lin$^{1*}$ \quad Hai Huang$^{1}$\quad Hanting Wang$^{1}$\quad Sihang Cai$^{1}$ \quad WenKang Han$^{1}$ \\  Tao Jin$^{1}$\quad Jingyuan Chen$^{1}$ \quad Jiacheng Sun$^{2}$ \quad Jieming Zhu$^{2}$ \quad Zhou Zhao$^{1\dagger}$
\\
$^{1}$ Zhejiang University\qquad
$^{2}$ Huawei Noah’s Ark Lab \\
{\tt\small shuleiwang@zju.edu.cn},\quad
{\tt\small zhaozhou@zju.edu.cn}
}

\usepackage[misc]{ifsym}
\newcommand\blfootnote[1]{
    \begingroup
    \renewcommand\thefootnote{}\footnote{#1}
    \addtocounter{footnote}{-1}
    \endgroup
}
\usepackage[hang]{footmisc}

\begin{document}
\twocolumn[{
    \maketitle
    \vspace{-15pt}
    \includegraphics[width=\linewidth]{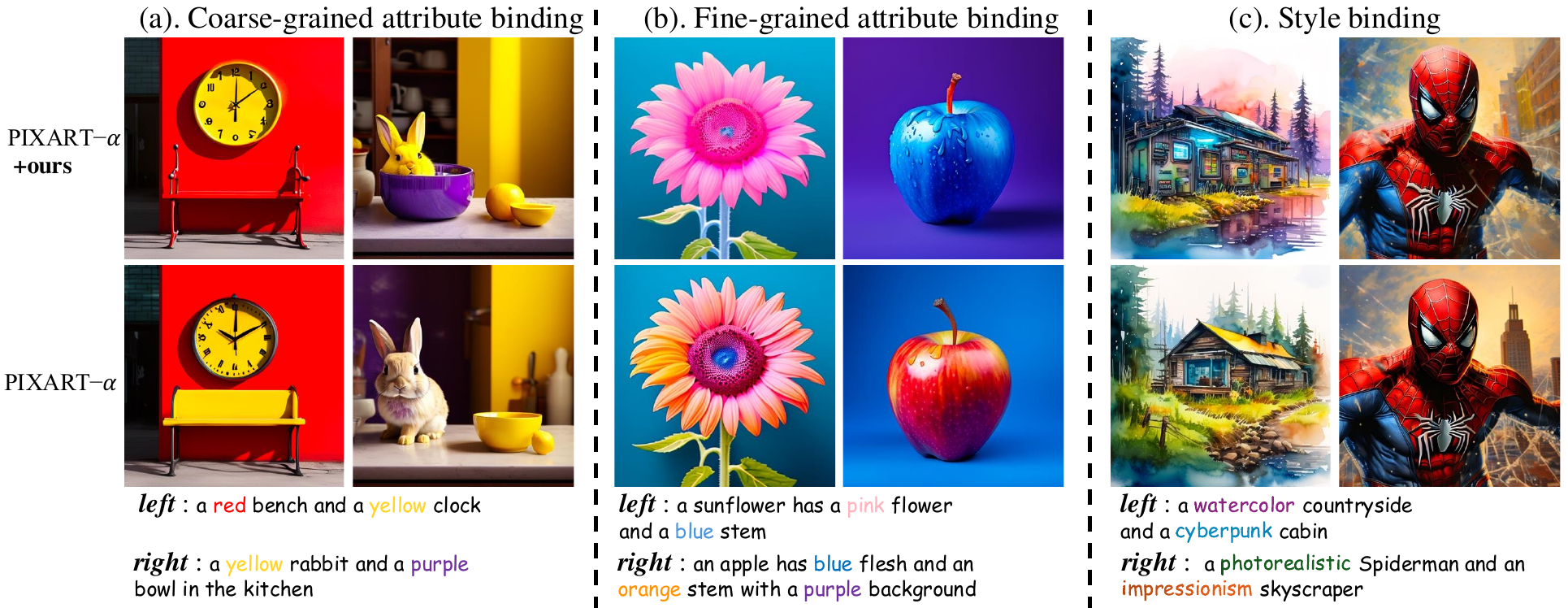}
    \captionof{figure}{Our method directly optimizes the cross-attention maps in Transformer-based diffusion models, significantly enhancing the model's performance in coarse-grained attribute binding and further improving fine-grained attribute and style binding. For instance, our approach enables precise control over the color of an apple’s flesh and stem as well as the style of two distinct concepts.}\vspace{0.3cm}
    \label{fig:teaser}
}]

\maketitle

{
    \blfootnote{
        $^*$Equal contribution.\\
        $^\dagger$Corresponding author\\
        }
}

\begin{abstract}
We introduce a novel, training-free approach for enhancing alignment in Transformer-based Text-Guided Diffusion Models (TGDMs). Existing TGDMs often struggle to generate semantically aligned images, particularly when dealing with complex text prompts or multi-concept attribute binding challenges.
Previous U-Net-based methods primarily optimized the latent space, but their direct application to Transformer-based architectures has shown limited effectiveness. Our method addresses these challenges by directly optimizing cross-attention maps during the generation process. Specifically, we introduce Self-Coherence Guidance, a method that dynamically refines attention maps using masks derived from previous denoising steps, ensuring precise alignment without additional training. 
To validate our approach, we constructed more challenging benchmarks for evaluating coarse-grained attribute binding, fine-grained attribute binding, and style binding. Experimental results demonstrate the superior performance of our method, significantly surpassing other state-of-the-art methods across all evaluated tasks. Our code is available at {\tt\small \href{https://scg-diffusion.github.io/scg-diffusion}{https://scg-diffusion.github.io/scg-diffusion}}.
\end{abstract}    
\vspace{-10pt}
\section{Introduction}
\label{sec:intro}

Text-guided diffusion models (TGDMs)~\cite{rombach2022high, saharia2022photorealistic, ramesh2022hierarchical} have demonstrated significant capability, generating high-quality images based on given textual prompts. 
Previous mainstream TGDMs~\cite{rombach2022high, saharia2022photorealistic} have primarily utilized the U-Net architecture~\cite{ronneberger2015u}, predicting noise through down-sampling and up-sampling modules. Recently, Transformer~\cite{vaswani2017attention}, due to its robustness~\cite{wang2021pyramid}, scalability~\cite{peebles2023scalable}, and efficient modality fusion~\cite{girdhar2023imagebind,ji2024wavtokenizer,huang2024autogeo}, has been widely applied in large language models(LLMs)~\cite{ouyang2022training, achiam2023gpt, touvron2023llama} and multi-modal large language models(MLLMs)~\cite{liu2024visual, zhu2023minigpt, li2023blip}.

Peebles \textit{et al.}~\cite{peebles2023scalable} were the first to explore the use of transformers in generative tasks.This pioneering work was followed by several subsequent works~\cite{chen2023pixart,li2024hunyuan,chen2023gentron} that have enabled high-quality image generation and demonstrated strong scalability.




Aligned generation is a challenging task that requires the generated images and text to maintain a high level of alignment~\cite{xia2024achieving,huang2024semantic,huang2024unlocking}. Nonetheless, we observe that the existing TGDMs, whether based on U-Net or Transformer architectures, struggle to maintain precise alignment, especially for complex prompts. We categorize these semantic discrepancies as follows:

\textbf{a)} Coarse-grained attribute binding: For instance, as shown in Fig.~\ref{fig:teaser}(a), when prompted with ``a red bench and an yellow clock", the model may incorrectly assign attributes to concepts, potentially mixing attributes between them.

\textbf{b)} Fine-grained attribute binding: When referring to a specific part of a concept, the model may misattribute details. For example,  as shown in Fig.~\ref{fig:teaser}(b), if prompted with ``an apple with a blue fruit and an orange stem", the model may fail to apply the colors correctly to the intended parts.
\textbf{c)} Style binding: When required to generate multiple concepts in distinct styles, as shown in Fig.~\ref{fig:teaser}(c), the model may incorrectly assign styles to concepts, leading to misaligned visual outcomes.


Previous work has explored aligned generation based on U-Net models, but these methods show limited effectiveness when applied to Transformer-based models, as shown in Fig.\ref{fig:direct_apply}. Therefore, we revisited the differences between the U-Net and Transformer architectures, as illustrated in Fig.\ref{fig:U-Net_dit}. Our findings indicate that in Transformer-based architectures, it is more challenging to identify attention maps with strong core semantic information to guide the latent generation process.

Therefore, we delve deeper into the Transformer-based framework to address these three challenges. Upon closely analyzing the Transformer-based architecture, We find that, due to the absence of downsampling and upsampling modules in the Transformer architecture, the shape of the cross-attention maps during the generation process remains unchanged. This property provides us with the convenience of directly operating the cross-attention maps. Can we directly optimize the maps during the generation process in the same way as the cross attention editing~\cite{hertz2022prompt}? The answer is no. This is because the cross attention editing is designed for image editing task, where an existing generated image guides the creation of another image. In contrast, direct text-to-image generation relies solely on text as guidance, without an additional image to guide the process. So we propose a training-free, self-coherence guidance approach for cross-attention maps. Specifically, we dynamically extract masks from the previous denoising step’s features, then apply these masks to refine the current step’s attention map for enhanced alignment.



To validate the effectiveness of our method, we constructed a more challenging benchmark for fine-grained attribute and style binding, building upon previous attribute-binding benchmarks. We conducted comprehensive evaluations of the generated results, including qualitative and quantitative assessments, as well as a user study. To summary, our main contributions are as follows:
\begin{itemize}
    \item We propose a novel, training-free approach for transformer-based TGDMs that directly optimizes the cross attention map rather than the latent representation. 
    \item To validate our approach, we introduce more challenging benchmarks, evaluating not only simple coarse-grained attribute binding but also fine-grained attribute binding and style binding.
    \item Extensive experiments verify the effectiveness of our method, demonstrating superior performance over baselines and state-of-the-art U-Net-based methods.
\end{itemize}
\section{Related work}
\label{sec:Related work}

\subsection{Text-Guided Diffusion Model}

TGDMs~\cite{rombach2022high,saharia2022photorealistic}, such as Stable Diffusion(SD)~\cite{rombach2022high} and Imagen~\cite{saharia2022photorealistic}, have demonstrated remarkable capabilities by generating high-quality images from given textual prompts. Current mainstream TGDMs generally fall into two categories: those based on U-Net architectures and those based on Transformer architectures. Previous SOTA models have primarily utilized the U-Net-based approach, as seen in models like Stable Diffusion~\cite{rombach2022high}. However, the successful application of transformers in large language models and multimodal models has highlighted their significant advantages. Recently, Peebles \textit{et al.}~\cite{peebles2023scalable} proposed DiT, a fully Transformer-based architecture, which showcases greater scalability. PixArt-$\alpha$~\cite{chen2023pixart} further advances this approach by incorporating a multi-head cross-attention mechanism that bridges the text and image modalities, allowing high-quality image generation with reduced training data requirements.

\subsection{Aliged generation}

Despite the remarkable success of diffusion models in text-to-image generation, even state-of-the-art models, such as Stable Diffusion, struggle to generate images that highly align with text prompts~\cite{tang2022daam, feng2022training, chefer2023attend, wang2022diffusiondb,lin2024non,linaction}, particularly when dealing with complex semantics or attribute binding. Consequently, a range of methods~\cite{hu2024ella, agarwal2023star, zhu2024isolated, bao2024separate, rassin2024linguistic, kim2023dense, liang2024rich, wen2023improving} has been proposed to improve alignment in pre-trained diffusion T2I models. Current approaches can be categorized into two main types: those requiring fine-tuning and those that are training-free.
\begin{figure*}[t]
    \centering
    \includegraphics[width=0.9\linewidth]{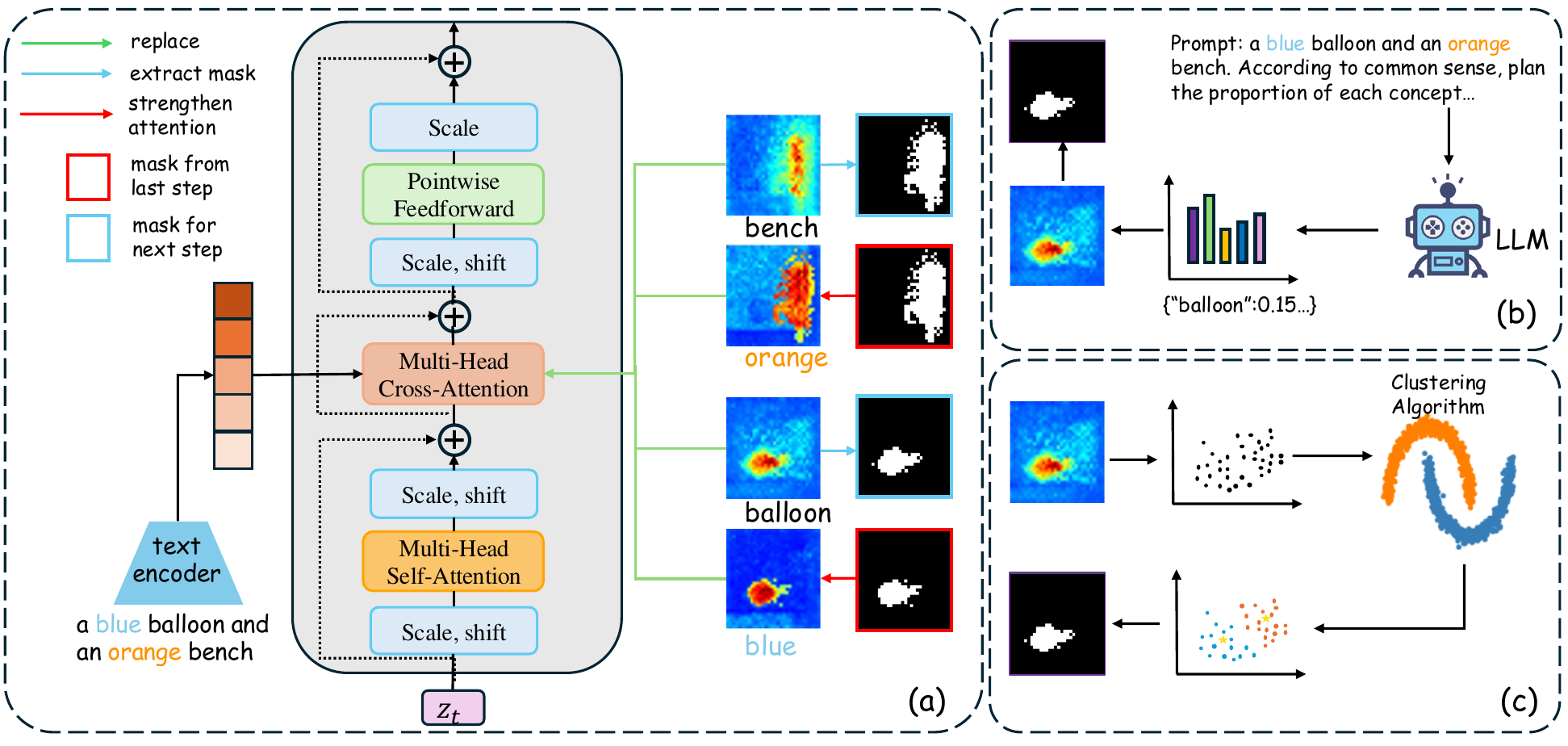}
    \caption{ \textbf{(a)} Overview of our method. Given a prompt, we extract the corresponding concept masks and use these masks to directly guide the attribute or style maps. \textbf{(b)} For fine-grained attribute binding, we extract masks by planning the proportions using LLMs. \textbf{(c)} For coarse-grained attribute binding and style binding, we directly apply clustering methods to extract the corresponding masks.}
    \label{fig:methods}
    \vspace{-1em}
\end{figure*}

\textbf{Fine-tuning Methods}: Methods requiring fine-tuning often involved lightweight adjustments or multimodal alignment to enhance control over generation. For instance, Mou \textit{et al.}~\cite{mou2024t2i} introduced a lightweight adapter that allowed for rich control of the generated results. Feng~\textit{et al.}~\cite{feng2024ranni} incorporated a semantic panel to decouple T2I tasks into "text-to-semantic panel" and "semantic panel-to-image" processes and constructed a relevant dataset for model fine-tuning, thereby achieving aliged generation. Shen~\textit{et al.}~\cite{shen2024sg} employed a scene graph adapter to deconstruct text into scene graphs, enabling SD to capture semantic relationships~\cite{lin2023tavt,lin2023exploring} more effectively.

\textbf{Training-Free Methods}: Training-free approaches~\cite{chefer2023attend, zhang2024spdiffusion, guo2024initno, Li2023DivideB, meral2024conform, zhuang2024magnet} typically optimized the inference process directly, avoiding the computational demands and potential issues like catastrophic forgetting associated with fine-tuning. Chefer \textit{et al.}~\cite{chefer2023attend} introduced Generative Semantic Nursing, using features from the generation process to guide latent generation by activating concept tokens in the cross-attention map; however, it did not fully resolve attribute binding issues. Li \textit{et al.}~\cite{Li2023DivideB} further designed a binding loss to align color and concept cross-attention maps more closely. Meral \textit{et al.}~\cite{meral2024conform} leveraged contrastive learning to achieve promising results by pushing apart mismatched attribute-concept pairs and pulling closer matched attribute-concept pairs.


While these methods have improved generation alignment to some extent, they remain limited in handling fine-grained attribute and style bindings. Notably, previous training-free approaches optimize the latent space, whereas our method directly optimizes the cross-attention map. This not only improves coarse-grained attribute binding but also enables precise fine-grained attribute and style binding.

\section{Methodology}
In this section, we begin by introducing the concept of TGDMs and denoising network, followed by a detailed presentation of our proposed method for directly optimizing the attention map. The framework of our approach is illustrated in Fig.~\ref{fig:methods}

\begin{figure*}[ht]
    \centering
    \includegraphics[width=\linewidth]{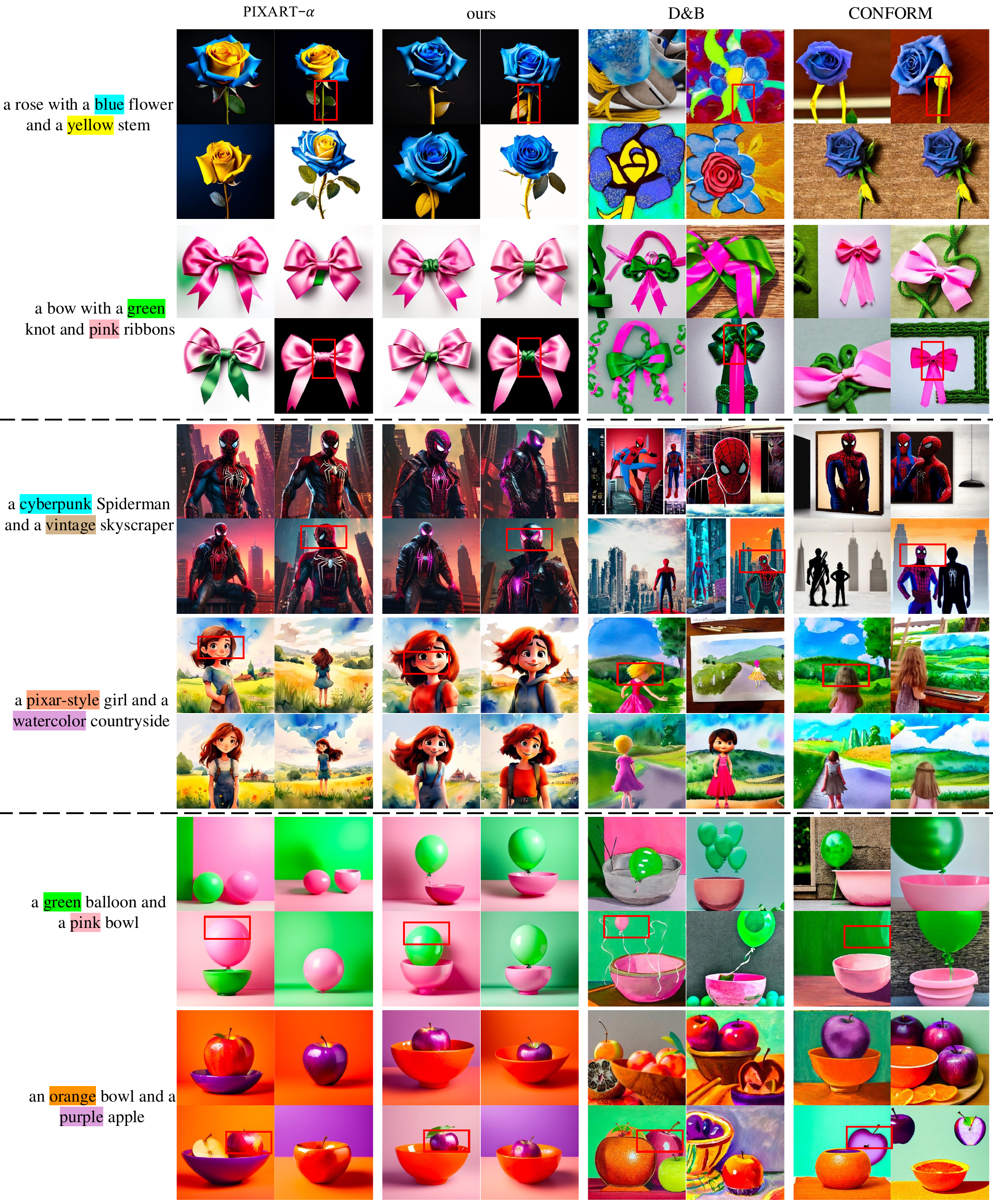}
    \caption{Qualitative analysis of our method compared to other SOTA methods.Our approach consistently generates high-quality images with superior alignment across coarse-grained attribute binding, fine-grained attribute binding, and style binding tasks.}
    \label{fig:main_exp}
    \vspace{-1em}
\end{figure*}
\subsection{Preliminaries}
\paragraph{Text-Guided Diffusion Models}
Our method is primarily based on the text-guided diffusion model. Existing approaches usually operate predominantly in the latent space rather than at the pixel level. Specifically, a Variational Autoencoder(VAE) includes an encoder $\mathcal{E}$ that encodes an input image $x$ into a lower-dimensional latent code $z$
, and a decoder $\mathcal{D}$ that reconstructs 
$z$ back into the original image $\mathcal{D}(z) \approx x$.
Normally, the VAE remains fixed during the training of the denoising network .

In detail, noise $\epsilon$ is gradually added to the initial latent $z_{0}$ to produce a noisy latent representation $z_{t}$, and a denoising network $\epsilon_\theta$ is trained to predict the noise added to $z_{0}$. The specific training objective is formulated as follows:

\vspace{-10pt}
\[
\mathcal{L}=\mathbb{E}_{z \sim \mathcal{E}(x), \epsilon \sim N(0, I), c, t}\left[\left\|\epsilon-\epsilon_\theta\left(z_t, c, t\right)\right\|^2\right] \tag{1}
\]
Here, $c$ denotes conditional information, specifically referring to the text input.
During inference, we begin by randomly sampling $z_{t}$ from a Gaussian distribution. The denoising network iteratively estimates and removes noise from $z_{t}$ based on both $z_{t}$ and the condition $c$, ultimately obtaining the clean latent code $z_{0}$. By operating within the latent space, this approach enables more efficient and robust image synthesis.

\paragraph{Denoising Network $\epsilon_\theta$ in TGDMs}

Current denoising networks can be broadly categorized into two types: U-Net-based and Transformer-based models. U-Net-based denoising networks leverage convolutional neural networks (CNNs), while Transformer-based models are entirely built upon attention mechanisms. In Transformer-based TGDMs architectures, there are typically $n$ Transformer blocks, each consisting of a self-attention module and a cross-attention module.
In the self-attention module, latent features are encoded as $Q$, $K$, and $V$. The attention weights are computed using the following equation: 
$\text{Attention}(Q, K, V) = \text{softmax}\left(\frac{QK^\top}{\sqrt{d_k}}\right)V$. 
In the cross-attention module, text embeddings—typically generated by a frozen text encoder~\cite{raffel2020exploring, radford2021learning}—are encoded as $K$, and $V$ vectors by the cross-attention layer, enabling the integration of text-based conditional information.while the hidden states from the transformer block are projected to generate the $Q$ vectors. Using the formula 
$A_{t}=\operatorname{Softmax}\left(\frac{Q K^T}{\sqrt{d}}\right)$, where $t$ denotes the time step, we obtain the cross-attention map. The softmax operation is performed along the final dimension. To more intuitively represent the semantic information of the cross-attention map, we typically reshape $A_{t}$ into dimensions 
$\mathbb{R}^{h \times w \times L}$, where 
$h, w$ denote the spatial dimensions of the map, and $l$ represents the length of the text sequence. The attention map for the $s$-th token is denoted as $A_{t}^{s}$.It is noted that both U-Net-based models and Transformer-based models have cross-attention maps, and Prompt-to-Prompt~\cite{hertz2022prompt} demonstrates that cross-attention maps contains rich semantic information, which plays a critical role in determining the content of the generated image.





\subsection{Self-Coherence Guidance}

Our purpose is to enhance the performance of Transformer-based TGDMs in aligned generation. To achieve this, we first compare the differences between Transformer-based and U-Net-based architectures. Intuitively, there is a significant structural difference: the Transformer architecture does not include upsampling and downsampling modules. As a result, the hidden state shape remains constant across all layers, and the shape of the cross-attention map remains unchanged.

Upon deeper investigation, we observe differences in their behavior during the generation process. As illustrated in Fig.~\ref{fig:U-Net_dit}, the 
Transformer-based TGDMs exhibit more evenly distributed semantic information in the cross-attention map during generation, making it more difficult to identify the core map.

Inspired by the cross attention editing~\cite{hertz2022prompt}, which has shown impressive results in image editing tasks, we recognize that this mechanism requires a reference image's cross-attention map as guidance. However, our task involves direct text-to-image generation without a reference image. Therefore, based on the above analysis, we propose the Self-Coherence Guidance, with the core idea being to directly optimize the maps during the generation process.


Specifically, TGDMs can generate different parts of two concepts or a single concept. It knows    ``what to draw``, but not ``where to draw" it~\cite{tewel2023key,patashnik2023localizing}, leading to errors in attribute binding.
For instance, when the prompt is ``a blue balloon and an orange bench" or ``an apple with blue flesh and a red stem", TGDMs can generally generate complete parts of the balloon, bench, and apple. However, it often assigns incorrect colors to the corresponding parts.

Therefore, we can leverage the prior knowledge that TGDM knows ``what to draw" to implement self-coherence guidance.

We denote concept tokens as 
$o$ and attribute or style tokens as 
$r$ at step $t$, where 
$z_{t}$
generates 
$z_{t-1}$
, we leverage the concept tokens at step 
$t+1$
 (such as balloon, stem, etc.) to extract a mask 
$M_{t+1}^{o_{i}}$and directly enhance $A_{t}$.To achieve this enhancement, we define the following update rule for $\widehat{A_{t}}$:

\vspace{-10pt}
{\small
\[
\left(\widehat{A_{t}}\right)_{p, q}:= \begin{cases} 
c \cdot\left(A_{t}\right)_{p, q} & \text{if } q = r_i \text{ and } M_{t+1}^{o_i}[p] == 1, \\ 
\left(A_{t}\right)_{p, q} & \text{otherwise.} 
\end{cases} \tag{2} \label{eq:scg}
\]
}


Here, 
$p$
represents a position index within the cross-attention map, $q$ denotes a token index, and $i$ indicates the index of an concept-attribute pair.

\begin{figure}[t]
    \centering
    \includegraphics[width=0.9\linewidth]{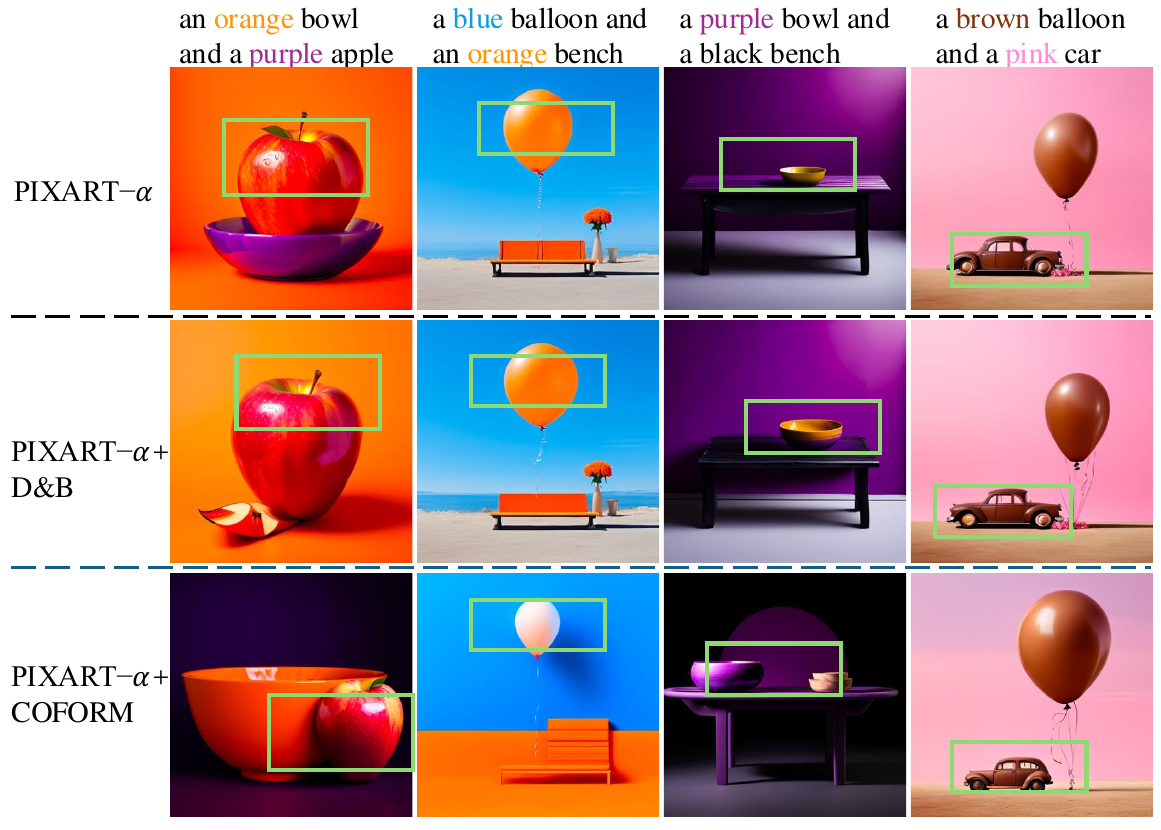}
    \caption{Qualitative results of directly transferring D\&B and CONFORM methods to Transformer-based architectures.}
    \label{fig:direct_apply}
    \vspace{-1em}
\end{figure}

To obtain $M_{t+1}^{o_{i}}$,we first average the attention maps across all layers at step $t+1$ and then adopt different methods depending on the specific task.
For coarse-grained attribute binding and style binding, we directly apply the K-means algorithm to the averaged cross-attention map, dividing it into two categories to generate the mask.
For fine-grained attribute binding, although the cross-attention map can still focus on the corresponding regions, clustering fails to effectively capture the detailed concepts. To address this, we leverage LLMs to determine the proportions. Specifically, given a prompt such as ``an apple with blue flesh and an orange stem", we query LLMs to leverage its commonsense reasoning and infer the approximate $ratio$ of the flesh and stem regions in an image of an apple. Based on the inferred $ratio$, we extract the mask by selecting regions in the cross-attention map whose values fall within the top proportion corresponding to the calculated $ratio$.

Then, we calculate $\widehat{A_{t}}$ using Equation (2) and replace the original 
$A_{t}$ with $\widehat{A_{t}}$ to directly guide the diffusion model in generating images that closely align with the text.

\section{Experiments}
Our method addresses challenges in Coarse-grained attribute binding, Fine-grained attribute binding, and Style binding in Transformer-based TGDMs. Our experiments are designed to address the following research questions:
\begin{itemize}
    \item \textbf{RQ1}: Can the previously proposed training-free disambiguation method based on U-Net be directly applied to Transformer-based models?

    \item \textbf{RQ2}: Can existing training-free disambiguation methods be extended to broader applications, such as fine-grained attribute binding and style binding?

    \item \textbf{RQ3}: Is Self-Coherence Guidance effective across all three scenarios?


\end{itemize}

\vspace{-15pt}
\paragraph{Experimental setup}
Existing benchmarks typically focus on evaluating coarse-grained attribute binding, but lack benchmarks and prompts for fine-grained attribute and style binding. For coarse-grained attribute binding, we adopt the prompts from prior work. However, upon analysis, we found that these structured prompts lack diversity and are not adapted to specific scenes. For example, they lack prompts like ``a [colorA] [animal] and a [colorB] [object]" or ``a [colorA] [animal] and a [colorB] [object] in the kitchen." Therefore, based on previous work~\cite{chefer2023attend}, we constructed a more challenging coarse-grained attribute binding benchmark using manually created prompts, enhancing prompt complexity. Additionally, we developed fine-grained attribute binding and style binding structured prompts for quantitative and qualitative analysis through a similar manual construction approach.

Specifically, the structure of the prompt for coarse-grained attribute binding is ``a [colorA] [objectA] and a [colorB] [objectB]", ``a [colorA] [animal] and a [colorB] [object]", ``a [colorA] [animal] and a [colorB] [object] in the [place]". For fine-grained attribute binding, the prompt structure is ``an object with a [colorA] [partA] and a [colorB] [partB]". For style binding, the prompt structure is ``a [styleA] [objectA] and a [styleB] [objectB]".The details of the benchmark can be found in the supplementary material.


We conduct experiments on PIXART-$\alpha$-512. For each prompt, we use 64 different random seeds with 50 iterations. The parameter $c$ is set to 4.
\begin{table}[t]
    \small
    \centering
    \setlength{\tabcolsep}{10pt}
    \caption{Comparison of image-text similarity and text-text similarity for the direct transfer of U-Net-based methods to Transformer-based models.
    \\[-0.7cm]} 
    \begin{tabular}{l c c} 
        \toprule
        Method & image-text & text-text  \\
        \midrule
        \textbf{PIXART-$\alpha$}~\cite{chen2023pixart}          & 0.36   & 0.807    \\
        \hspace{10pt}w/ D\&B~\cite{Li2023DivideB}    & 0.35   & 0.807      \\
        \hspace{10pt}w/ CONFORM~\cite{meral2024conform} & 0.36   & 0.814      \\
        \midrule
        \textbf{SD}~\cite{rombach2022high}         & 0.34  & 0.771\\
        \hspace{10pt}w/ D\&B~\cite{Li2023DivideB}         & 0.36  & 0.803\\
        \hspace{10pt}w/ CONFORM~\cite{meral2024conform}        & 0.36  & 0.824   \\
        
        \bottomrule \\[-0.8cm]
    \end{tabular}
    \label{tb:directly apply}
\end{table}
\vspace{-10pt}
\paragraph{Baselines}

To investigate the generation alignment challenges within Transformer-based architectures, we conduct experiments on the text-to-image model PIXART-$\alpha$~\cite{chen2023pixart}, which is based on the DiT architecture. Our baselines include the original PIXART-$\alpha$, Divide \& Bind (D\&B)~\cite{Li2023DivideB}, and CONFORM~\cite{meral2024conform}. It is worth noting that both D\&B and CONFORM are state-of-the-art training-free methods specifically designed for U-Net-based models. Initially, we simply adapt these U-Net-based SOTA methods to the transformer-based PIXART-$\alpha$ model. To further demonstrate the remarkable effectiveness of our method in addressing generation consistency and to explore the upper bounds of our method’s performance, we also compare it with the original U-Net-based SOTA methods, all of which are based on Stable Diffusion v1.5 and have demonstrated strong capabilities in various scenarios.




\begin{figure}[t]
    \centering
    \includegraphics[width=\linewidth]{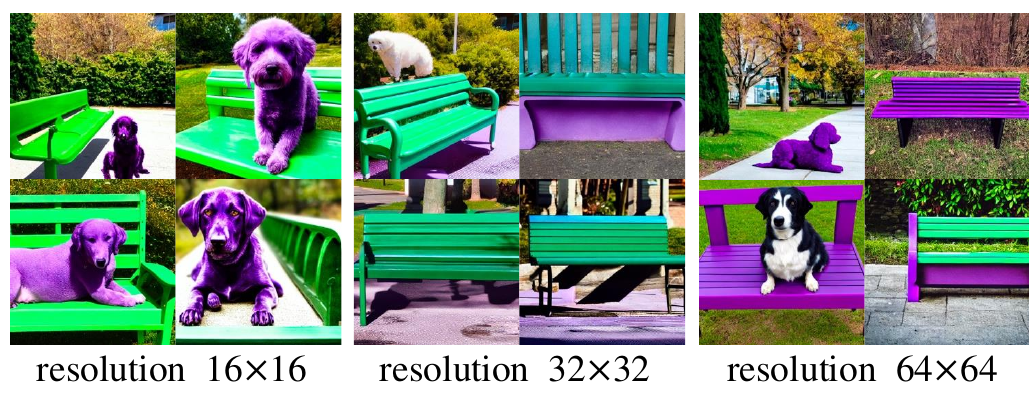}
    \caption{Generation results of U-Net-based CONFORM using cross-attention maps at different resolutions,where the results at corresponding positions are generated using the same random seed.The text prompt is ``a purple dog and a green bench".}
    \label{fig:resolution}
    \vspace{-1em}
\end{figure}
\subsection{Direct Transfer of U-Net-Based Methods to Transformer-Based model(RQ1)}

We begin by directly applying the D\&B~\cite{Li2023DivideB} and CONFORM~\cite{meral2024conform} methods to PIXART-$\alpha$, following the original parameter configurations to conduct experiments on a original objects binding benchmark~\cite{chefer2023attend}.
\vspace{-10pt}
\paragraph{Qualitative Results}

Figure \ref{fig:direct_apply} illustrates the experimental results of PIXART-$\alpha$ and the direct application of D\&B and CONFORM to PIXART-$\alpha$, with each method using the same seed. Neither D\&B nor CONFORM effectively addresses the attribute binding issues in Transformer-based architectures. We summarize the challenges of directly transferring U-Net-based methods into three main areas. (a): Difficulty in achieving full attribute binding, often resulting in only one attribute being correctly bound. For instance, when the prompt is ``an orange bowl and a purple apple", CONFORM correctly binds the color orange to the bowl but fails to bind purple to the apple. (b): Reduced generation quality. As shown in the Fig.~\ref{fig:direct_apply}, both D\&B and CONFORM can result in incomplete or malformed concepts. (c): Potential for decreased semantic alignment. For example, when the prompt is ``a purple bowl and a black bench", PIXART-$\alpha$ successfully generates a bowl and a bench, but CONFORM generates an additional bowl, as illustrated in the Fig.~\ref{fig:direct_apply}.
\begin{figure}[t]
    \centering
    \includegraphics[width=\linewidth]{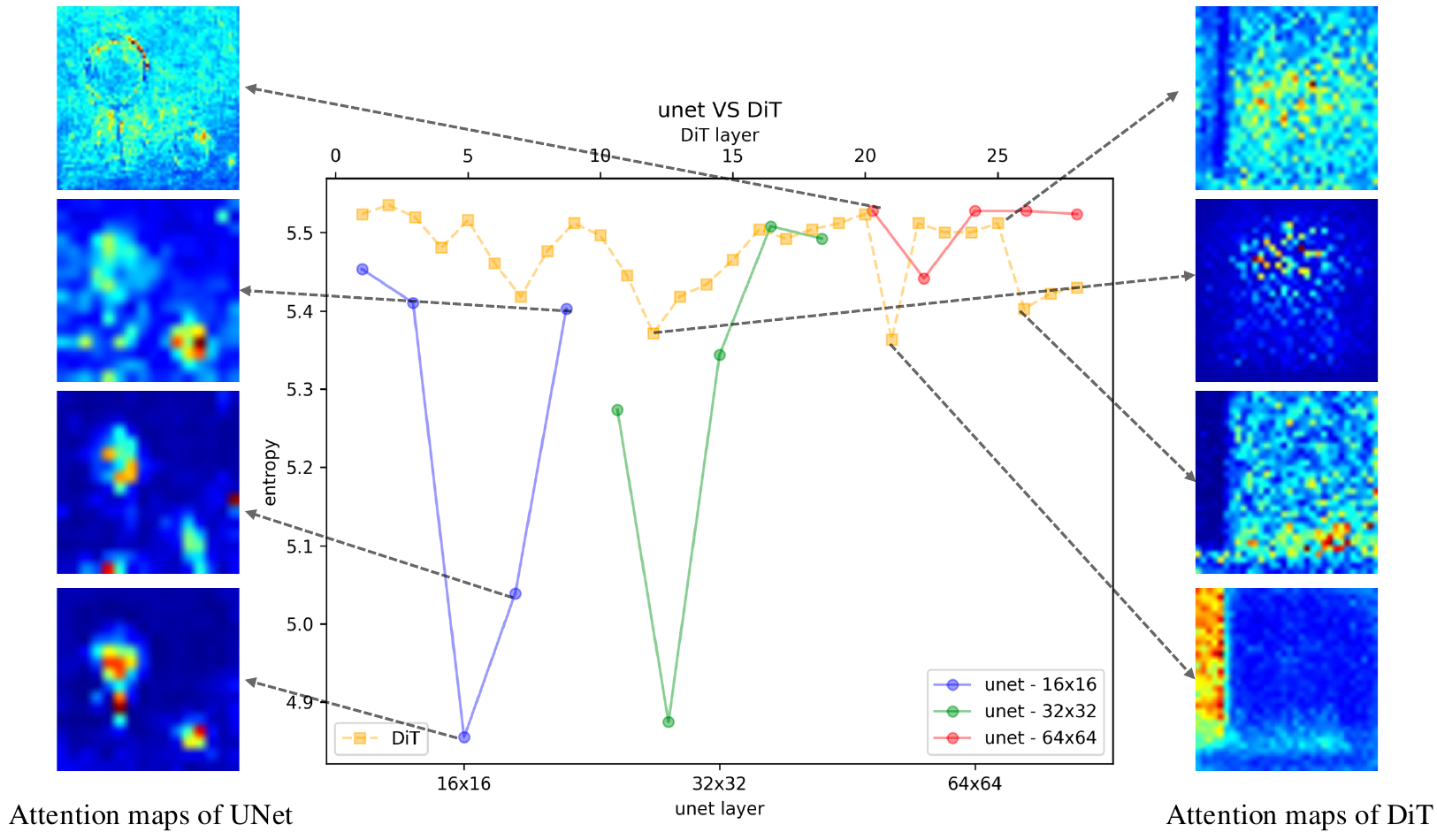}
    \caption{During the generation process, the attention entropy across different layers in U-Net and DiT architectures reflects the semantic richness, with lower attention entropy indicating greater semantic information.The visualized token corresponds to the word ``balloon".}
    \label{fig:U-Net_dit}
    \vspace{-1em}
\end{figure}
\vspace{-10pt}
\paragraph{Quantitative Results}
To further investigate the effects of directly applying D\&B and CONFORM, we conducted a quantitative analysis of the generated results using several metrics. We assessed image-text similarity and text-text similarity. Following previous methodologies, we used CLIP~\cite{radford2021learning} to separately encode the image and text, then calculated their similarity. Additionally, we employed the BLIP model~\cite{li2022blip} to generate captions for the images, subsequently calculating the similarity between the generated caption and the text prompt.

As shown in the table \ref{tb:directly apply}, while the use of the CONFORM method improves text-text similarity compared to PIXART-$\alpha$, it still lags behind the performance of the SD-based Conform approach. This suggests that directly applying U-Net-based methods yields limited effectiveness.

The limited effectiveness of directly applying U-Net-based methods to Transformer architectures motivates us to explore the underlying causes. Through an analysis of both U-Net and Transformer architectures, we have identified the following insights: due to the presence of downsampling and upsampling modules within the U-Net architecture, it produces cross-attention maps at varying resolutions. Among these, the 16x16 resolution maps are particularly notable for capturing richer semantic information~\cite{meral2024conform, hertz2022prompt}. Consequently, many U-Net-based methods tend to leverage the 16x16 cross-attention map.

To further verify whether the semantic information at 16x16 is indeed stronger, we set the resolution in CONFORM to 16, 32, and 64, respectively. As shown in the Figure \ref{fig:resolution}, generation quality is superior when the resolution is set to 16, while it significantly degrades at 32 and 64, indicating that the 16x16 cross-attention map does contain richer semantic information. To quantitatively compare the semantic information in U-Net and Transformer, we visualized the entropy~\cite{attanasio2022entropy} of the cross-attention maps for each layer. Lower entropy reflects stronger semantic information. As illustrated in the Figure \ref{fig:U-Net_dit}, the overall semantic strength is higher in U-Net’s 16x16 maps during generation. Although entropy decreases slightly in the final layers of DiT, the trend remains relatively stable throughout.

When directly adapting U-Net-based methods, the inherent characteristics of the Transformer—where all cross-attention maps share the same resolution and semantic information is more evenly distributed throughout the generation process—limit our ability to identify a core map with distinctly stronger semantic features.

Moreover, U-Net-based SOTA methods extract features via cross-attention and apply loss functions (e.g., contrastive or JS divergence) to guide latents, introducing an inherent gap that is aggravated by the relatively low semantic consistency of Transformer-based attention maps.



\begin{table}[t]
    \small
    \centering
    \setlength{\tabcolsep}{3pt}
    \caption{Comparison of average text-text similarity, demonstrating that our method outperforms other baselines across all three subtasks.\colorbox{mycolor_green}{Green} represents the best results, while \colorbox{mycolor_blue}{blue} indicates the second-best results.}
    \begin{tabular}{l c c c} 
        \toprule
        Method & Coarse-grained & Fine-grained& Style \\
        \midrule
        D\&B(SD) \cite{Li2023DivideB}          & 0.798  & 0.742 & 0.660 \\
        CONFORM(SD) \cite{meral2024conform}       & \colorbox{mycolor_blue}{0.834}  & 0.774  & 0.642 \\
        PIXART-$\alpha$ \cite{chen2023pixart}        & 0.808   & \colorbox{mycolor_blue}{0.804}  & \colorbox{mycolor_blue}{0.688}   \\
        \midrule
        Ours          & \colorbox{mycolor_green}{0.852}   & \colorbox{mycolor_green}{0.827}  & \colorbox{mycolor_green}{0.689}   \\
        \bottomrule \\[-0.8cm]
    \end{tabular}
    \label{tb:blip-clip}
\end{table}

\begin{table}[t]
    \small
    \centering
    \setlength{\tabcolsep}{3pt}
    \caption{Comparison of BLIP-VQA scores. Our method also demonstrates state-of-the-art performance, significantly surpassing the original PIXART-$\alpha$ model in both coarse-grained and fine-grained attribute binding.
    \\[-0.7cm]} 
    \begin{tabular}{l c c c} 
        \toprule
        Method & Coarse-grained & Fine-grained& Style \\
        \midrule
        D\&B(SD) \cite{Li2023DivideB}           & 0.499  & 0.451 & 0.514 \\
        CONFORM(SD) \cite{meral2024conform}        & \colorbox{mycolor_blue}{0.667}  & \colorbox{mycolor_blue}{0.522}   & 0.498 \\
        PIXART-$\alpha$ \cite{chen2023pixart}        & 0.293   & 0.398  & \colorbox{mycolor_blue}{0.676}   \\
        \midrule
        Ours          & \colorbox{mycolor_green}{0.679}  & \colorbox{mycolor_green}{0.623}  & \colorbox{mycolor_green}{0.799}   \\
        \bottomrule \\[-0.8cm]
    \end{tabular}
    \label{tb:blip-vqa}
\end{table}

\subsection{Self-Coherence Guidance VS U-Net-based SOTA methods (RQ2, RQ3)}

The suboptimal outcomes observed when directly transferring U-Net-based methods to Transformer architectures have motivated us to seek a more effective solution. We propose a novel paradigm that leverages the prior knowledge of generative models to directly optimize cross-attention maps. This approach bypasses the inherent gap introduced by previous U-Net-based methods, which typically involve an intermediate step of extracting 
cross-attention map features, calculating a loss, and subsequently optimizing the latent space towards the desired direction. By avoiding this intermediate step, our method achieves improved results.
\vspace{-10pt}
\paragraph{Qualitative Results}
As shown in the Fig.~\ref{fig:main_exp}, we compare our results with the original PIXART-$\alpha$, SD-based D\&B, and SD-based CONFORM. All methods were evaluated using the same seed for consistency. We observe that: (a) In the traditional coarse-grained attribute binding task, our method not only generates images with high consistency with the text prompts but also achieves state-of-the-art generation coherence. Although CONFORM also demonstrates competitive consistency, it sacrifices image quality. In contrast, our method maintains high fidelity in the generated images while preserving alignment. (b) Previous methods have struggled to precisely control specific parts of a concept at a fine-grained level. For example, when given the prompt ``a sunflower with blue petals and a yellow stem", other methods either confuse the colors of different parts or separate the two parts of the concept, failing to maintain the intended semantics of the prompt. Our approach, however, successfully controls the color of each part, generating high-quality, highly aligned images. (c) In the style-binding task, we aim for the model to control the style of two distinct concepts independently. For instance, we may want Spider-Man in a realistic style and the background in an Impressionist style. We observe that previous methods fail to manage different styles for separate concepts, often resulting in subpar outputs. In contrast, our method achieves successful style binding.
\vspace{-10pt}
\paragraph{Quantitative Results}
We use text-text similarity and BLIP-VQA as evaluation metrics. When evaluating text-text similarity, we first generate captions for images using BLIP. However, the BLIP captioning model does not always describe detailed attributes of each concept~\cite{huang2023t2i}. Therefore, Huang~\textit{et al.}~\cite{huang2023t2i} proposes BLIP-VQA, which decouples complex text prompts into independent questions. Specifically, for an image generated from the prompt ``a purple dog and a green bench", two separate questions are asked: ``a purple dog?" and ``a green bench?"

As shown in the table \ref{tb:blip-clip}, our method consistently outperforms other baselines in text-text similarity evaluation. Although it achieves only a slight improvement over the original PIXART-$\alpha$ in style binding, we attribute this to a limitation in the BLIP-caption model: when generating captions, BLIP struggles to accurately capture the distinct styles of each concept, which reduces the text-text similarity score.

As shown in the table \ref{tb:blip-vqa}, in the BLIP-VQA evaluation, our method also surpasses the original PIXART-$\alpha$ and other U-Net-based SOTA methods across all three subsets. Notably, the original PIXART-$\alpha$ model performs worse than the U-Net-based SOTA methods in both coarse binding and fine binding tasks, further underscoring the necessity of exploring generation consistency within Transformer-based architectures. Our method significantly enhances the original PIXART-$\alpha$ model, achieving substantial improvements in coarse binding and a 56\% increase in fine binding, which strongly demonstrates the effectiveness of our approach. In style binding, while the original PIXART-$\alpha$ model also performs well, our method effectively leverages its prior knowledge, further extending its capabilities and achieving an 18\% improvement.
\begin{table}[t]
    \small
    \centering
    \setlength{\tabcolsep}{3pt}
    \caption{User study with 50 participants
    \\[0cm]} 
    \begin{tabular}{l c c c} 
        \toprule
        Method & Coarse-grained & Fine-grained& Style \\
        \midrule
        D\&B(SD) \cite{Li2023DivideB}           & 3.2\%  & 0.6\% & 1.4\% \\
        CONFORM(SD) \cite{meral2024conform}     & \colorbox{mycolor_blue}{13.2\%}  & \colorbox{mycolor_blue}{4.6\%}   & 0.4\% \\
        PIXART-$\alpha$ \cite{chen2023pixart}     & 2.6\%    & 4.2\%  & \colorbox{mycolor_blue}{5.6\%}   \\
        \midrule
        Ours          & \colorbox{mycolor_green}{81\%}   & \colorbox{mycolor_green}{90.6\%}  & \colorbox{mycolor_green}{92.6\%}    \\
        \bottomrule \\[-0.8cm]
    \end{tabular}
    \label{tb:user study}
\end{table}
\vspace{-10pt}
\paragraph{User study}
To further evaluate the generation quality of our method, we conducted a comprehensive user study involving 50 participants. Specifically, following the setup of previous methods \cite{chefer2023attend}, we randomly selected 10 prompts for each subtask and generated images using each method with the same random seed. Similarly, we asked participants to select the image that best matched the text prompt from those generated by different methods. We used the frequency of selection as the evaluation metric.
Table \ref{tb:user study} presents the results of our user study, showing that users consistently preferred the images generated by our method across all three tasks. This strong user preference highlights the effectiveness of our approach.


\vspace{-5pt}
\section{Conclusions}
In this work, we proposed a training-free Self-Coherence Guidance method to address alignment challenges in Transformer-based TGDMs. By directly optimizing cross-attention maps rather than latent spaces, our approach effectively improves coarse-grained, fine-grained, and style binding tasks, surpassing the performance of existing state-of-the-art methods. 
Our findings underscore the potential of leveraging attention map optimization as a pathway for addressing alignment issues, paving the way for future advancements in TGDMs.



{
    \small
    \bibliographystyle{ieeenat_fullname}
    \bibliography{main}
}

\clearpage
\setcounter{page}{1}
\maketitlesupplementary

\section{Benchmark Details}

\label{sec:Benchmark details}
To evaluate the capabilities of our model, we constructed a more comprehensive benchmark based on A\&E~\cite{chefer2023attend}. Through our analysis, we identified that previous benchmarks primarily focused on coarse-grained attribute binding and lacked specific and complex scenarios. For example, prior benchmarks often evaluated prompts such as "a purple dog and a green bench." To address this limitation, we augmented the coarse-grained attribute binding tasks with specific place. For example, our prompts include cases such as "a green rabbit and a yellow bowl in the kitchen."

We argue that coarse-grained attribute binding alone is insufficient to comprehensively evaluate model performance. Therefore, we further extended the benchmark with fine-grained attribute binding and style binding tasks. Specifically, we manually created 56 fine-grained attribute binding prompts and 48 style binding prompts. Fine-grained attribute binding requires the model to control the attributes of different parts of a concept, while style prompts demand that multiple concepts within a single image exhibit distinct styles, the style prompts include categories such as "cyberpunk," "watercolor," "photorealistic," "anime," and others. The details of our benchmark is presented in Table~\ref{tab:benchmark_sets}.

For quantitative evaluation, we used metrics including text-to-text similarity and BLIP-VQA and additionally employed image-text similarity evaluation as discussed in section~\ref{sec:More quantitative results}. Since both fine-grained attribute binding and style binding tasks involve only two concepts, we generated two questions per prompt for BLIP-VQA evaluation. For coarse-grained attribute binding tasks, as the generated image must adhere to specific locations, we generated three questions per prompt. For example, given the prompt "a blue dog and a red bench in the street," the corresponding questions are: "a blue dog?", "a red bench?", and "the street?"

This enhanced benchmark allows for a more comprehensive evaluation of model capabilities across coarse-grained, fine-grained, and style attribute bindings. Our evaluation was conducted on RTX 3090 GPUs, with each generation taking approximately 20 seconds.


\begin{table*}[h]
    \centering
    \setlength{\tabcolsep}{8pt}
    \caption{The details of our benchmark. BLIP-VQA refers to the number of generated questions when evaluated using the BLIP-VQA metric.} %
    \begin{tabular}{lccc}
        \toprule
        Task & Template \& Example  & Prompt number& BLIP-VQA \\
        \midrule
        \multirow{ 4}{*}{Coarse-grained} & a [colorA][conceptA] and a [colorB][conceptB] & \multirow{ 4}{*}{54}& \multirow{ 4}{*}{3}\\ 
        & \textit{`a \textbf{black backpack} and a \textbf{pink balloon'}} & & \\ \\
        &a [colorA][conceptA] and a [colorB][conceptB] in the [place]& &\\
        &\textit{`a \textbf{blue rabbit} and a \textbf{yellow bowl} in the \textbf{kitchen}'}& & \\ \\
        \\
        \multirow{2}{*}{Fine-grained} & a [concept] with a [colorA] [partA] and a [colorB][partB]& \multirow{ 2}{*}{56 }& \multirow{ 2}{*}{2}\\
        &\textit{`an apple with a \textbf{orange stem} and \textbf{blue flesh}'} & & \\
         \\
        \multirow{2}{*}{Style} & a [styleA][conceptA] and a [sytleB][conceptB]& \multirow{ 2}{*}{48 }& \multirow{ 2}{*}{2}\\
        & \textit{`a \textbf{anime cat} and a \textbf{photorealistic kitchen}'} & &\\
        \bottomrule \\[-0.8cm]
    \end{tabular}
    \label{tab:benchmark_sets}
\end{table*}

\begin{figure*}[t]
    \centering
    \includegraphics[width=\linewidth]{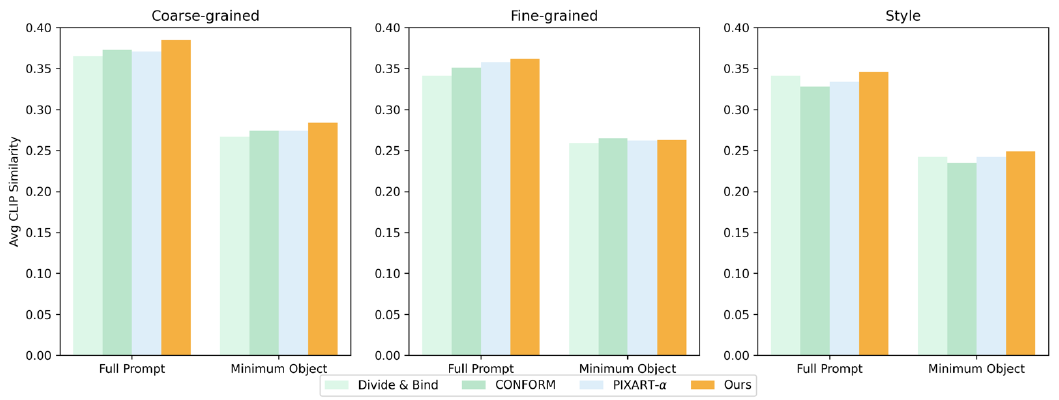}
    \caption{Comparison of average image-text similarity across different tasks. We compare our proposed method with the original PIXART-$\alpha$ model and two state-of-the-art aligned generation methods built on SD: D\&B and CONFORM.}
    \label{fig:clip}
    \vspace{-1em}
\end{figure*}


\section{Algorithm Details}
\label{sec:Algorithm}
\algnewcommand{\algorithmicgoto}{\textbf{Go to}}%
\algnewcommand{\Goto}[1]{\algorithmicgoto~\ref{#1}}%

\begin{algorithm}[t!]
\caption{Self-Coherence Guidance.}

\begin{flushleft}
\textbf{Input:} A text prompt $\mathcal{P}$, random seed $s$ and hyper-parameter $c$.\\
\textbf{Output:} The latent space $z_0$ corresponding to images with strong consistency to the text prompt $\mathcal{P}$.
\end{flushleft}

\begin{algorithmic}[1]
\For{$t=T,T-1,\ldots,1$}
    \State $z_{t-1}^*,A_t \gets DM(z_t,t)$  
    \State $M_t \gets Cluster(A_t) / LLMplanning(A_t)$ 
    \State $h^0=z_t$
    \For{$n=1,2,\ldots,N$}
        \State $A_{t}^n \gets TransBlock^n(h^{n-1})$
        \State $\widehat{A_{t}^n}\gets SCG(A_{t}^n,M_{t+1},c)$
        \State $h^n \gets TransBlock^n(h^{n-1})\{A_{t}^n \gets \widehat{A_{t}^n}\}$
   \EndFor
   \State $z_{t-1}=h^N$
\EndFor
\State \textbf{Return} $z_{0}$

\end{algorithmic}
\label{alg:scg}
\end{algorithm}
The process of our method is detailed in Algorithm~\ref{alg:scg}. Specifically, the $SCG$ function corresponds to the approach for obtaining the new attention map described in Equation~\ref{eq:scg}. Here, $N$ represents the number of Transformer blocks, $h$ represents the hidden state that each Transformer block outputs, $M_t$ represents the masks of the concepts extracted for the next step. We iteratively replace the original attention maps with the new attention maps for each block.

\section{More Quantitative Results}
\label{sec:More quantitative results}
To further quantitatively evaluate the performance of our method, we employed image-text similarity as a metric. Following~\cite{chefer2023attend,meral2024conform}, we utilized CLIP to separately encode images and their corresponding textual descriptions and computed their similarity scores, as shown in the Fig.~\ref{fig:clip}. In this evaluation, ``Full Prompts`` similarity refers to the similarity between the complete prompt and the image, while ``Minimum Object`` similarity measures the similarity between the image and the neglected half of the text prompt.

Our method consistently outperforms previous approaches across all three tasks, with significant improvements in average similarity for both coarse-grained and fine-grained attribute binding. Notably, while the CONFORM achieves results close to ours on coarse-grained attribute binding, it fails to generalize effectively to fine-grained attribute binding and style binding, showing the poorest performance in the latter. The original PIXART-$\alpha$ model performs reasonably well on fine-grained attribute binding, and our approach further enhances its performance, achieving the best results. However, for style binding, the improvement of our method over D\&B is relatively modest.

We attribute this limitation to the BLIP-caption model, which lacks specialized training for style-specific images. Consequently, it struggles to capture the fine-grained stylistic details of different concepts in images, demonstrating insensitivity to style. 

In addition, we also employed the evaluation metric VQAScore. VQAScore is similar to the BLIP-VQA metric, as both assess image-text alignment by leveraging a VQA model. As shown in table~\ref{tb:VQAScore}. 
Our method achieves SOTA results on this additional metric as well.

\begin{table}[t]
    \small
    \centering
    \setlength{\tabcolsep}{3pt}
    \caption{Comparison of VQAScore. Our method still achieves state-of-the-art performance on this metric. 
    \\[-0.7cm]} 
    \begin{tabular}{l c c c} 
        \toprule
        Method & Coarse-grained & Fine-grained& Style \\
        \midrule
        D\&B(SD) \cite{Li2023DivideB}           & 0.372  & 0.342 & 0.307 \\
        CONFORM(SD) \cite{meral2024conform}        & \colorbox{mycolor_blue}{ 0.412}  & 0.376   & 0.312 \\
        PIXART-$\alpha$ \cite{chen2023pixart}        & 0.387   &  \colorbox{mycolor_blue}{0.437}  & \colorbox{mycolor_blue}{ 0.320}   \\
        \midrule
        Ours          & \colorbox{mycolor_green}{0.476}  & \colorbox{mycolor_green}{0.488}  & \colorbox{mycolor_green}{0.366}   \\
        \bottomrule \\[-0.8cm]
    \end{tabular}
    \label{tb:VQAScore}
\end{table}

To achieve a more accurate analysis, we provide a more comprehensive qualitative evaluation in the following section.

\section{Ablation Study}
We conduct ablation studies to compare the performance of LLM and K-means approaches. Specifically, we evaluate both methods across three tasks: coarse-grained attribute binding, fine-grained attribute binding, and style binding. BLIP-VQA is used as the evaluation metric. The results are shown in table~\ref{tb:ablation}.
Our experimental results demonstrate that the LLM-based approach performs better on the fine-grained attribute binding task.We attribute this to the following three reasons.
\textbf{First}, the saliency of attention maps varies across different tasks, with fine-grained tasks exhibiting the least salient attention maps. \textbf{Second}, clustering algorithms process signals directly from the model output, whereas LLMs incorporate external knowledge to assist in interpreting the output. As a result, when the attention map is highly salient, clustering algorithms achieve better performance.
\textbf{Finally}, LLMs provide additional benefits only when the attention map lacks saliency, making them more effective in fine-grained tasks.
\begin{table}[h]
    \small
    \centering
    \setlength{\tabcolsep}{8pt} 
    \renewcommand{\arraystretch}{1.2} 
    \caption{Ablation study comparing different grouping strategies.}
    \begin{tabular}{lccc}
        \toprule
        Method & Coarse-grained & Fine-grained & Style \\
        \midrule
        w/ LLM    & 0.647 & \textbf{0.623} & 0.781 \\
        w/ K-means & \textbf{0.679} & 0.613 & \textbf{0.799} \\
        \bottomrule
    \end{tabular}
    \label{tb:ablation}
\end{table}

\section{Generalizability}
\begin{figure}[t]
    \centering
    \includegraphics[width=\linewidth]{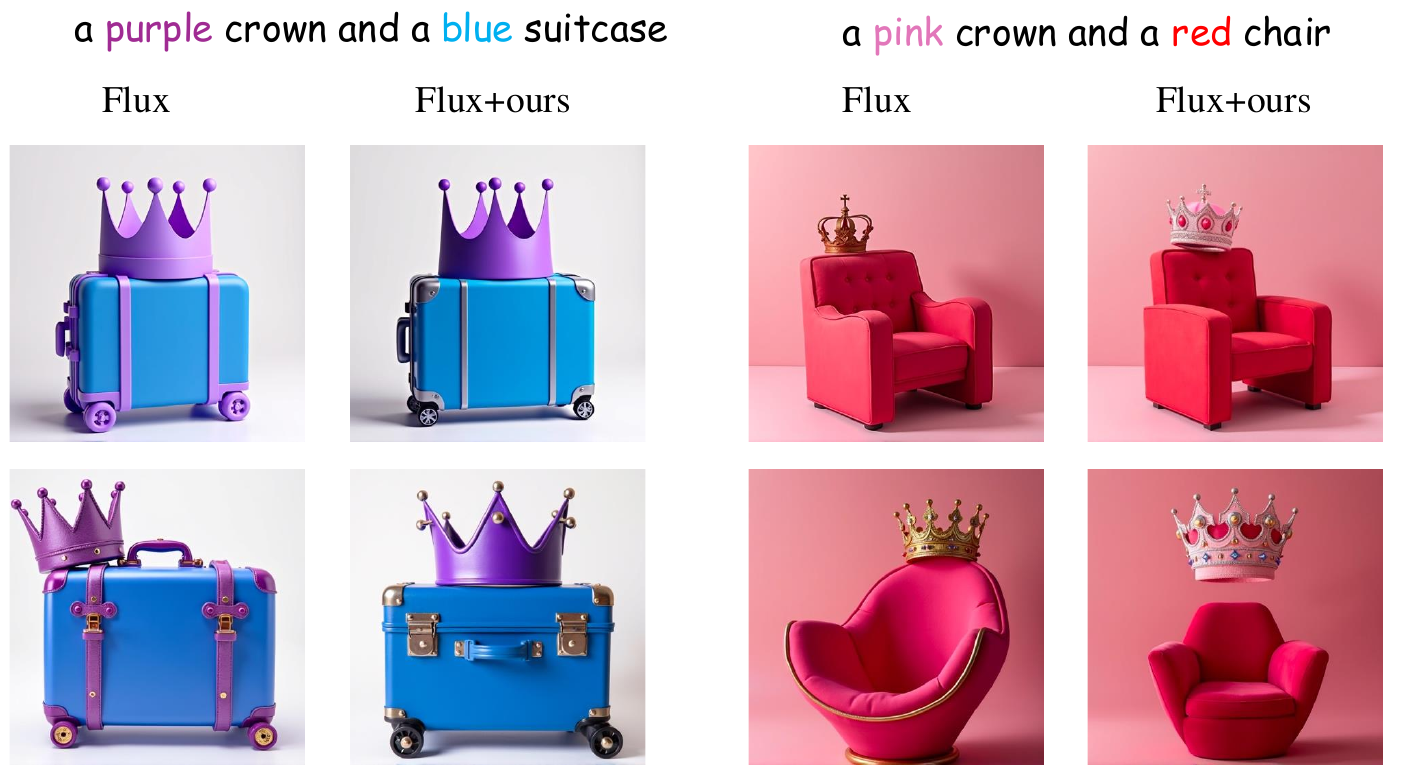}
    \caption{The experimental results on Flux demonstrate that our method remains effective under the MMDiT architecture..}
    \label{fig:flux-exp}
    \vspace{-1em}
\end{figure}
To verify the generalization capability of our method, we further conduct experiments on Flux~\cite{flux}. Unlike PIXART-$\alpha$~\cite{chen2023pixart}, Flux~\cite{flux} adopts the MMDiT architecture, which concatenates the QKV of text and image modalities before computing attention. In this setting, we still treat the dot product between the image queries and text keys as the cross-attention map on which our method operates. As shown in Fig~\ref{fig:flux-exp}, our approach remains effective under the MMDiT architecture, demonstrating strong generalization ability.

\section{More Qualitative Results}
\label{sec:More qualitative results}
To further validate the effectiveness of our approach, we provide additional qualitative analysis results for fine-grained attribute binding, style binding, and coarse-grained attribute binding tasks.These results further validate the effectiveness of our method.

\begin{figure*}[ht]
    \centering
    \includegraphics[width=\linewidth]{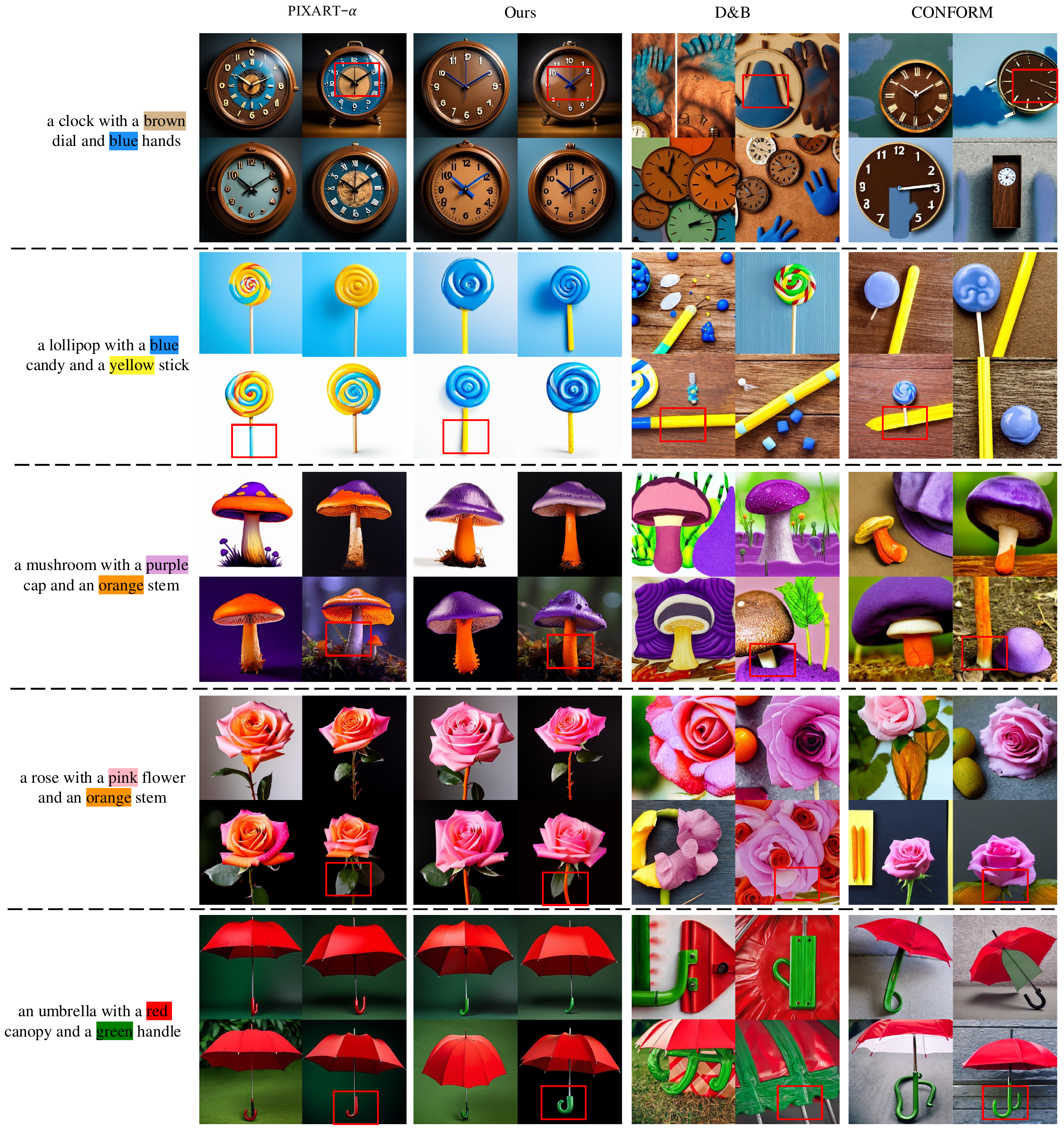}
    \caption{Qualitative analysis of fine-grained attribute binding comparing our method with other SOTA approaches. Our method enables more precise control over the attributes of concepts.}
    \label{fig:qual_1}
    \vspace{-1em}
\end{figure*}

\begin{figure*}[ht]
    \centering
    \includegraphics[width=\linewidth]{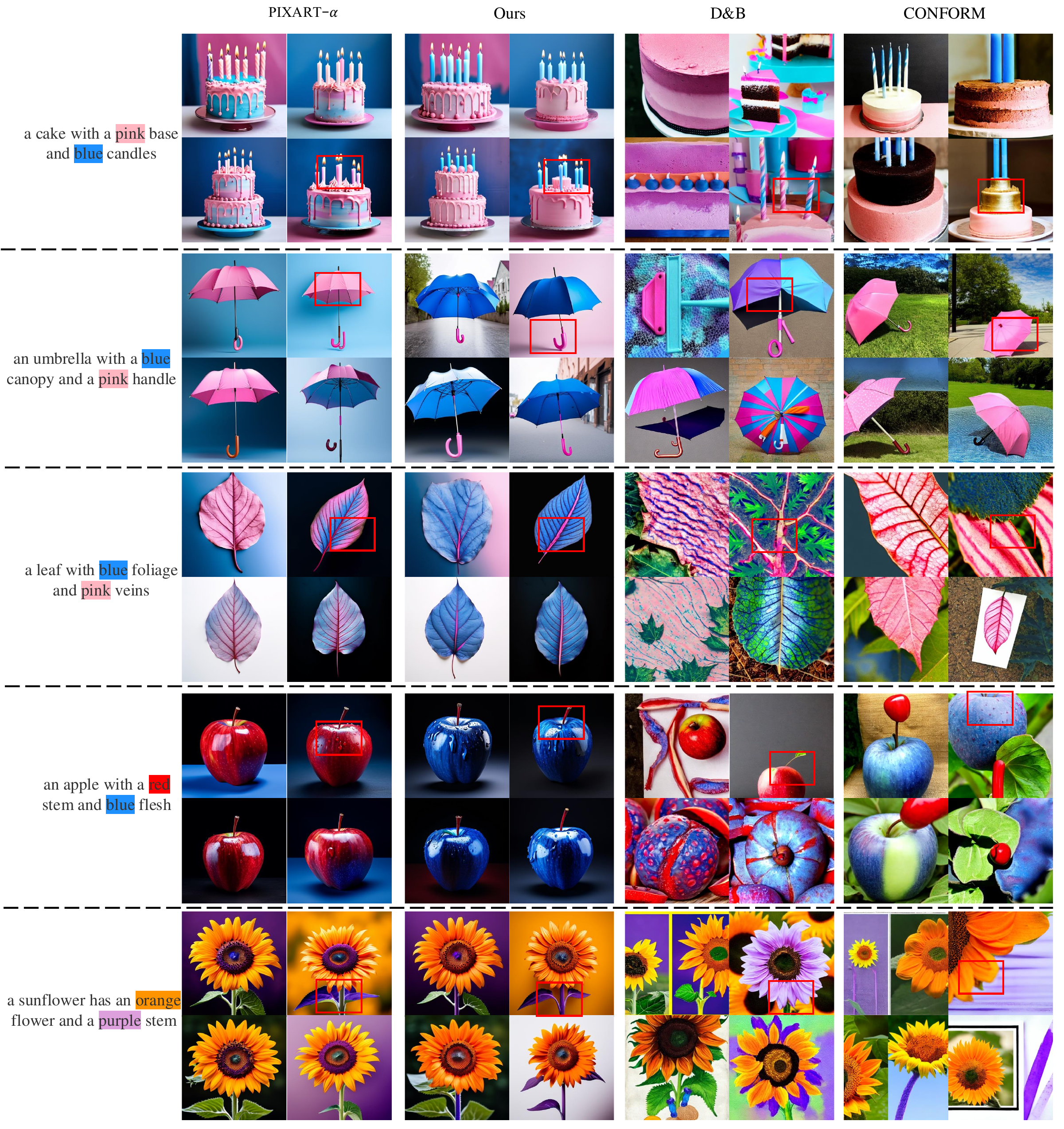}
    \caption{Qualitative analysis of fine-grained attribute binding comparing our method with other SOTA approaches. Our method enables more precise control over the attributes of concepts.}
    \label{fig:qual_2}
    \vspace{-1em}
\end{figure*}

\begin{figure*}[ht]
    \centering
    \includegraphics[width=\linewidth]{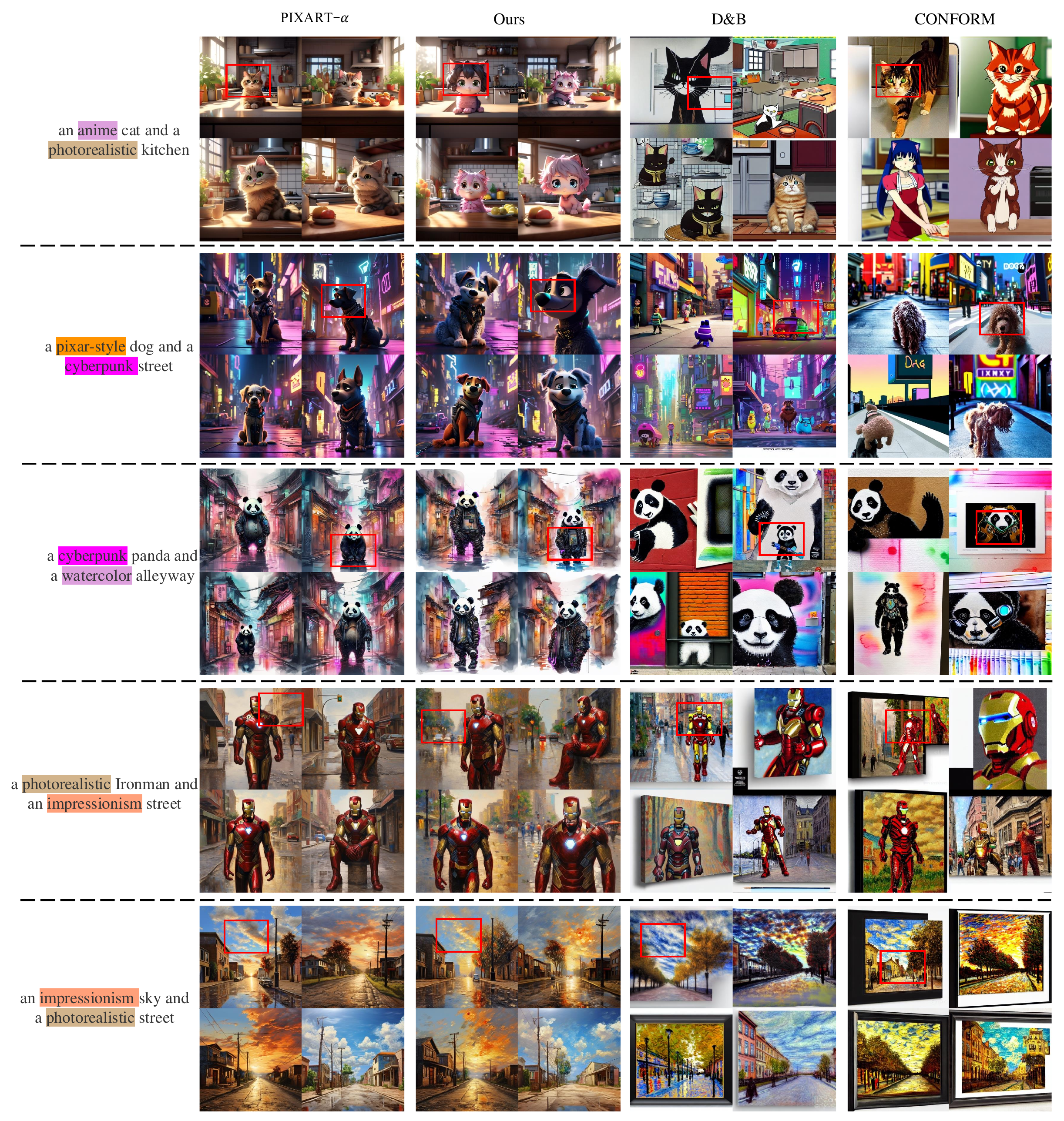}
    \caption{Qualitative analysis of style binding comparing our method with other SOTA approaches. Our method effectively binds styles to different concepts.}
    \label{fig:qual_3}
    \vspace{-1em}
\end{figure*}

\begin{figure*}[ht]
    \centering
    \includegraphics[width=\linewidth]{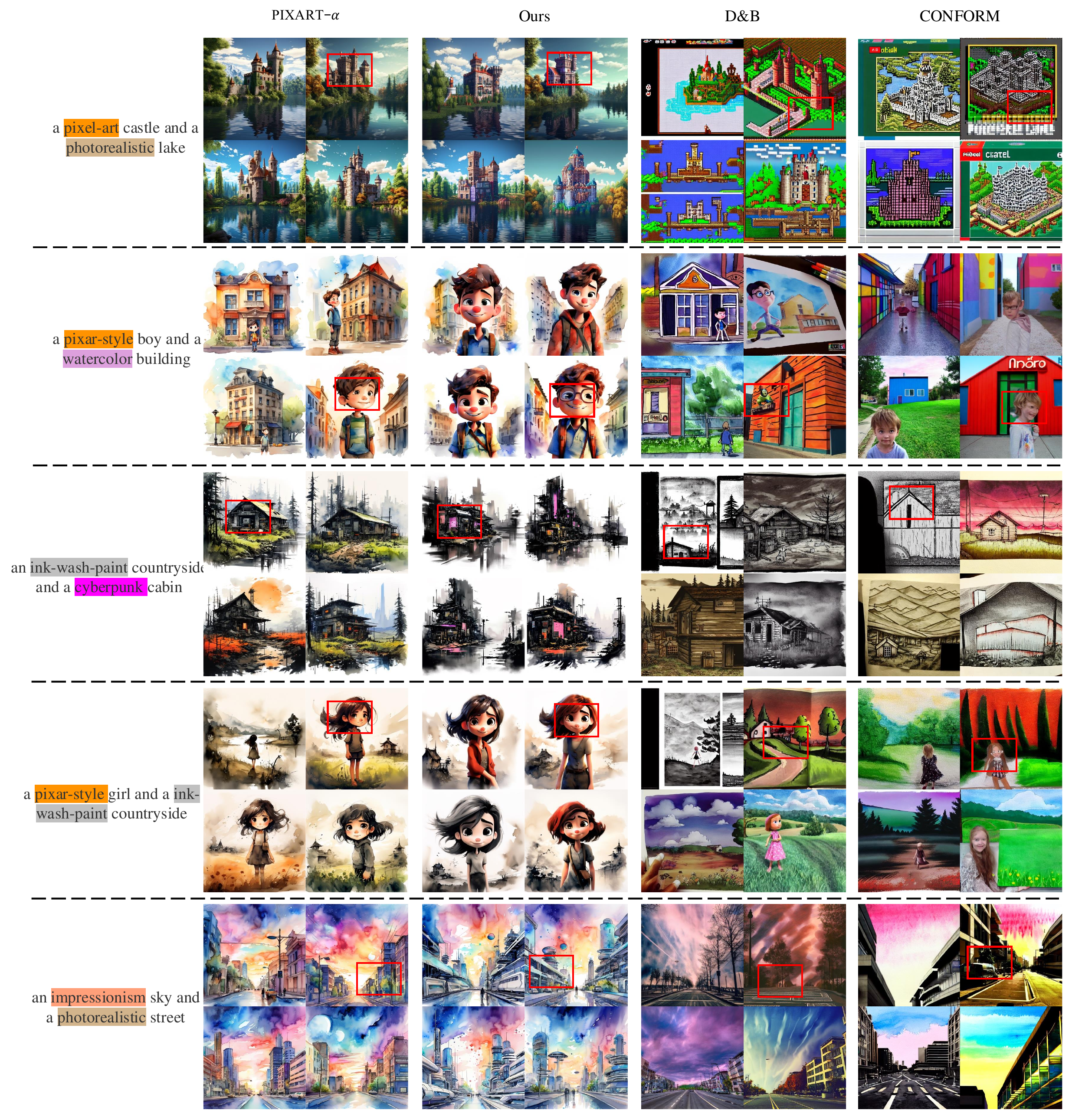}
    \caption{Qualitative analysis of style binding comparing our method with other SOTA approaches. Our method effectively binds styles to different concepts.}
    \label{fig:qual_4}
    \vspace{-1em}
\end{figure*}

\begin{figure*}[ht]
    \centering
    \includegraphics[width=\linewidth]{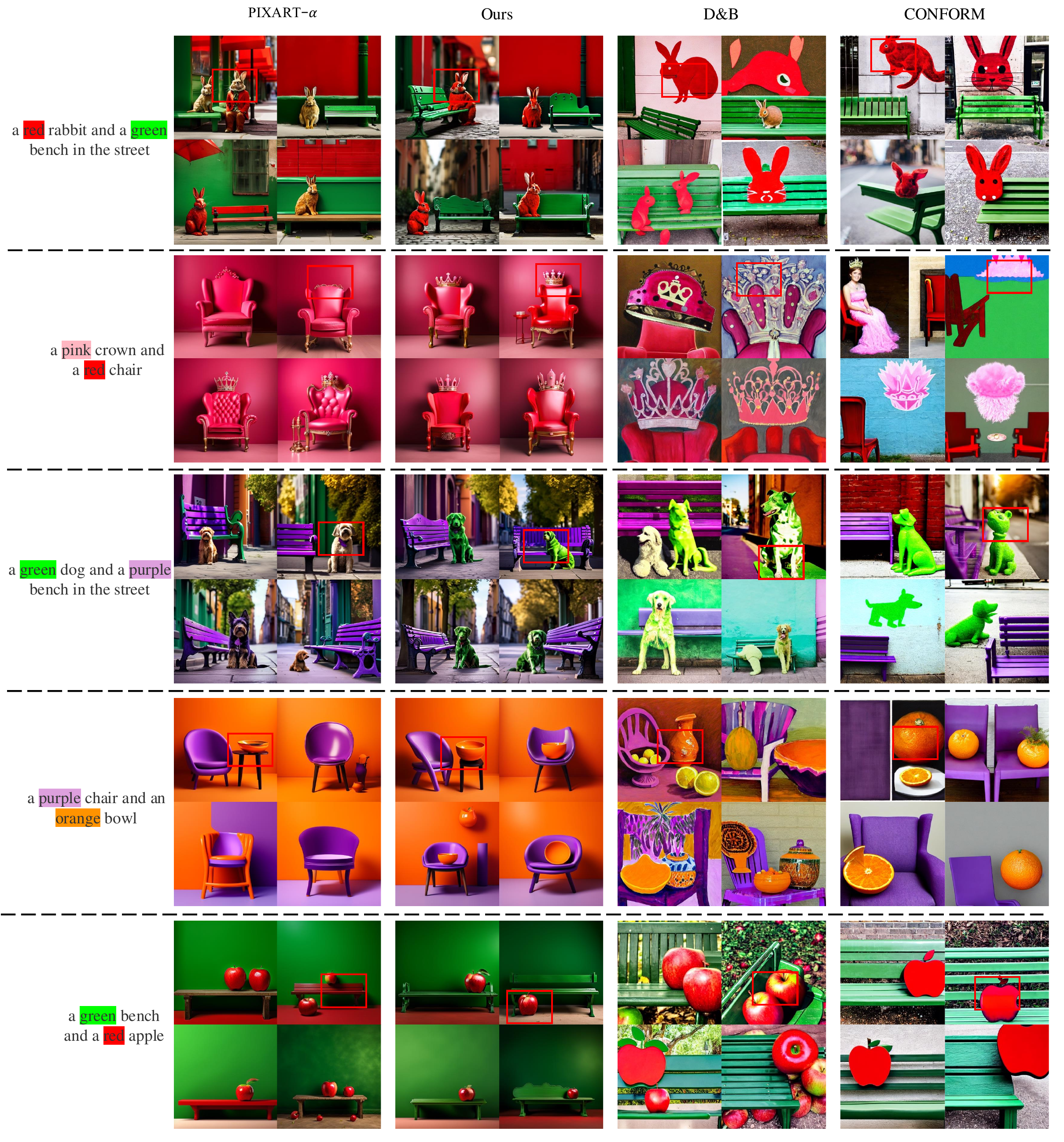}
    \caption{Qualitative analysis of coarse-grained attribute binding comparing our method with other SOTA approaches. Our method not only achieves attribute control but also generates higher-quality concepts.}
    \label{fig:qual_5}
    \vspace{-1em}
\end{figure*}

\begin{figure*}[ht]
    \centering
    \includegraphics[width=\linewidth]{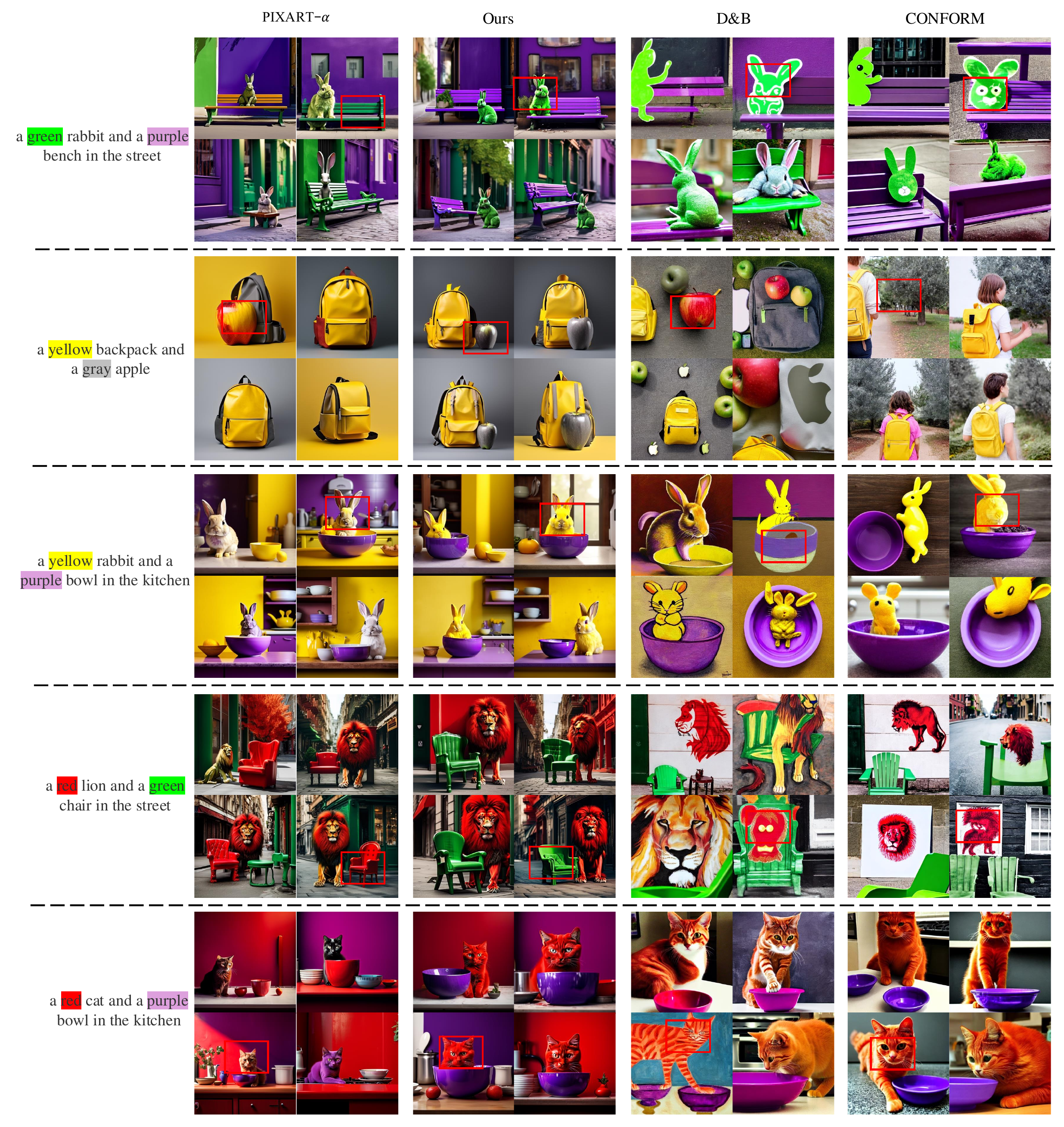}
    \caption{Qualitative analysis of coarse-grained attribute binding comparing our method with other SOTA approaches. Our method not only achieves attribute control but also generates higher-quality concepts.}
    \label{fig:qual_6}
    \vspace{-1em}
\end{figure*}


\end{document}